\documentclass[review]{elsarticle}
\usepackage{subfigure}
\usepackage{booktabs}
\usepackage{graphicx}
\usepackage{multirow}

\journal{Neurocomputing}

\newcommand{\Rmnum}[1]{\uppercase\expandafter{\romannumeral #1}}

\begin{document}

\begin{frontmatter}

\title{MMFNet: A Multi-modality MRI Fusion Network for Segmentation of Nasopharyngeal Carcinoma}
\author[mymainaddress]{Huai Chen}

\author[mymainaddress]{Yuxiao Qi}
\author[mysecondaryaddress]{Yong Yin}
\author[mysecondaryaddress]{Tengxiang Li}
\author[mythirdaddress]{Xiaoqing Liu}
\author[mythirdaddress]{Xiuli Li}
\author[mysecondaryaddress]{Guanzhong Gong\corref{mycorrespondingauthor}}
\cortext[mycorrespondingauthor]{Corresponding author}
\ead{gongguanzhong@yeah.net}
\author[mymainaddress]{Lisheng Wang\corref{mycorrespondingauthor}}
\cortext[mycorrespondingauthor]{Corresponding author}
\ead{lswang@sjtu.edu.cn}
\address[mymainaddress]{Institute of Image Processing and Pattern Recognition, Department of Automation, Shanghai Jiao Tong University, Shanghai, 200240, P. R. China}
\address[mysecondaryaddress]{Shandong Cancer Hospital Affiliated to Shandong University, Jinan, 250117, P.R China}
\address[mythirdaddress]{Deepwise AI Lab, P.R China}

\begin{abstract}
Segmentation of nasopharyngeal carcinoma (NPC) from Magnetic Resonance Images (MRI) is a crucial prerequisite for NPC radiotherapy. However, manually segmenting of NPC is time-consuming and labor-intensive. Additionally, single-modality MRI generally cannot provide enough information for its accurate delineation. Therefore, a multi-modality MRI fusion network (MMFNet), which is a novel framework to fuse information from multi-modality medical images, is proposed to utilize MRI of T1, T2 and contrast-enhanced T1 to complete accurate segmentation of NPC. The backbone of MMFNet is designed as a multi-encoder-based network, consisting of several encoders to capture modality-specific features and one decoder to obtain fused features for NPC segmentation. A fusion block is presented to effectively fuse multi-source features. It contains a 3D Convolutional Block Attention Module (3D-CBAM), recalibrating low-level features captured from modality-specific encoders to highlight both informative features and regions of interest (ROIs), and a residual fusion block (RFBlock), which fuses re-weighted features to keep balance between fused ones and high-level features from decoder. Moreover, in order to make full mining of individual information from multi-modality MRI, a training strategy named self-transfer is proposed to utilize pre-trained modality-specific encoders to initialize multi-encoder-based network. The proposed method based on multi-modality MRI can effectively segment NPC and its advantages are validated by extensive experiments.
\end{abstract}


\end{frontmatter}

\section{Introduction}
\label{section:introduction}
Nasopharyngeal carcinoma (NPC), which has an unknown and complicated etiology, is a kind of malignant tumor. The distinctive geographic distribution of NPC makes some regions such as Southeast Asia, South China, the Arctic and the Middle East/North Africa have extremely higher incidence than other regions \cite{mohammed2017review}. Patients with early detection and diagnosis of NPC will have a greater 10-year survival rate with 98\% for stage \Rmnum{1} and 60\% for stage \Rmnum{2}, while the median survival of patients at advanced stage is only 3 years \cite{wu2018nasopharyngeal}. Therefore, timely and effective treatments play a crucial role in reducing the mortality of NPC. Radiotherapy, which highly depends on medical images such as the Magnetic Resonance Images (MRI) to get accurate delineation of the gross tumor volume (GTV) to separate normal adjacent tissues from lesion regions to reduce radiation-associated toxicity \cite{razek2012mri}, is the mainstay of treatment for NPC. However, clinicians need to manually mark the boundary of NPC slice by slice before developing radiotherapy plans currently, which is time-consuming and labor-intensive. Additionally, the quality of manual segmentation highly depends on the experience of clinicians, which influences treatment effect. Therefore, an automatic and accurate segmentation approach is urgently needed to alleviate the workload of clinicians and improve the efficacy of treatment presently.

Recently, according to the type of features used, methods of NPC segmentation can be divided into two categories, one is based on traditional handcrafted features, and the other one is based on deep features obtained from deep neural network (DNN). In traditional methods, besides relatively simple manual features of medical images, such as intensity \cite{tatanun2010automatic,chanapai2012nasopharyngeal}, texture \cite{zhou2003texture} and shape \cite{fitton2011semi,huang2015nasopharyngeal}, some traditional knowledge-based methods such as support vector machine (SVM) \cite{zhou2011segmentation,huang2013region}, semi-supervised fuzzy c-means \cite{zhou2002mri,zhou2003texture}, dictionary learning \cite{wang2018tumor} are also implemented to generate NPC boundary. Huang et al. \cite{huang2015nasopharyngeal} proposed a three-step NPC segmentation method in MRI. In this method, an adaptive algorithm and a distance regularized level set were applied to obtain the region of NPC and the contour, then a novel HMRF-EM framework based on the maximum entropy is presented to further refine segmentation results. Wang et al. \cite{wang2018tumor} introduced a joint dictionary learning methods, which obtains simultaneously multiple dictionaries of CT, MRI and corresponding label, to achieve NPC segmentation.

Frameworks based on handcrafted features and traditional knowledge-based methods have been successfully implemented in the aforementioned papers to complete NPC segmentation. Nevertheless, the complex anatomical structure of NPC and the similarity of intensities between nearby tissues make it difficult to be accurately segmented only with the help of manual features. Meanwhile, the high diversity of shapes and sizes makes this task more challenging \cite{huang2015nasopharyngeal}. Therefore, inspired by the success of deep learning technology, some methods \cite{ma2018discriminative,ma2018automated,wang2018automatic} based on DNN were proposed to get more accurate segmentation in recent years. Ma et al. \cite{ma2018discriminative} developed an image-patch-based convolutional neural network (CNN), integrating two CNN-based classification networks into a Siamese-like sub-network, to combine CT and T1-weighted (T1) images to complete NPC segmentation. Ma et al. \cite{ma2018automated} proposed a method combining CNN and graph cut. According to this framework, the initial segmentation was firstly generated through integrating results obtained from three CNN-based networks focusing on axial, sagittal and coronal view. Then, a 3D graph-cut-based method was utilized to further refine it.

\begin{figure*}[]
	\centering	
	\subfigure[]{
		\begin{minipage}[b]{0.3\textwidth}
			\centering
			\includegraphics[width=\textwidth]{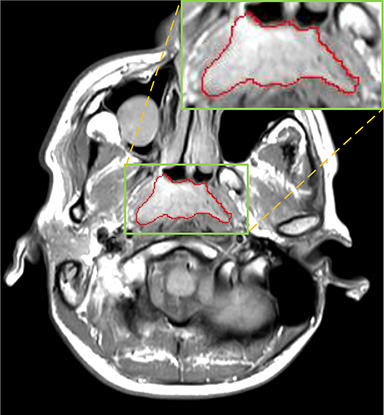}
		\end{minipage}
		\label{fig:multi_modalities:T1}
	}
	\subfigure[]{
		\begin{minipage}[b]{0.3\textwidth}
			\centering
			\includegraphics[width=\textwidth]{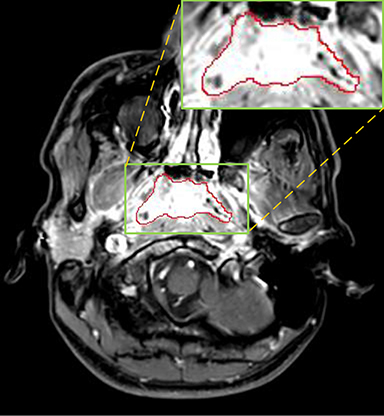}
		\end{minipage}
		\label{fig:multi_modalities:CET1}
	}
	\subfigure[]{
		\begin{minipage}[b]{0.3\textwidth}
			\centering
			\includegraphics[width=\textwidth]{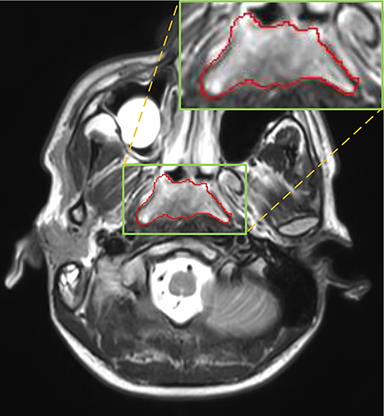}
		\end{minipage}
		\label{fig:multi_modalities:T2}
	}
	\caption{Examples of slices from different MRI (T1, CET1 and T2), the contour of NPC is marked in red line. (a),(b) and (c) are slices from T1 , CET1 and T2 respectively.}
	\label{fig:multi_modalities}
\end{figure*}

Previous deep-learning-based researches have established some excellent frameworks for NPC segmentation. Nevertheless, there remain the following deficiencies.
\begin{itemize}
	\item[(1)] Although some DNN-based frameworks have been proposed recently to improve the performance of NPC segmentation \cite{ma2018discriminative,ma2018automated,wang2018automatic}, all of them are patch-based methods, which use a fixed-scale sliding window to crop image and determine the class of the central point. The performance of these frameworks highly depends on the scale of sliding window and is dramatically time-consuming due to severe computing redundancy. 
	\item[(2)] All of above methods make predictions based on 2D slices of medical images, ignoring the vital role of 3D information in decision-making. Although Ma et al. \cite{ma2018automated} propose a method utilizing information of three-views to make decisions, this framework still fails to make full use of 3D information. 
	\item[(3)] Currently, there is still no effort to fuse multi-modality MRI to develop an automatic segmentation system for NPC. According to researches of Popovtzer et al. \cite{popovtzer2014mri}, it should be a routine clinical practice to incorporate all kinds of MRI datasets in highly conformal radiation therapy to realize GTV delineation of NPC. For delineation of NPC, MRI is the perferred imaging modality for its superior soft tissue contrast \cite{popovtzer2014mri,razek2012mri}. Moreover, MRI of different modalities data have different visual characteristics and various responses to different tissues and anatomical structures. For example, T1-weighted (T1) MRI is suitable for detecting skull base involvement and fat planes, while contrast-enhanced T1-weighted (CET1) MRI is used to identity tumor extent \cite{razek2012mri}. Figure \ref{fig:multi_modalities} shows some examples of NPC response in T1, CET1 and T2-weighted (T2) MRI.
\end{itemize}

In this paper, we develop a multi-modality MRI fusion network (MMFNet), which is a novel framework to effectively captain interdependencies of multi-source features from 3D medical images, to improve NPC segmentation by fusing multi-modality MRI (T1, CET1 and T2). In the MMFNet, the backbone, which contains several encoders and one single decoder, can be used to well learn both modality-specific and fused features used implicitly for NPC segmentation in each modality of MRI. We propose a fusion block to fuse modality-specific features. It can be divided into a 3D Convolutional Block Attention Module (3D-CBAM), which is an attention module for 3D medical images and recalibrates multi-source features to highlight informative features and the regions of interest (ROIs), and a residual fusion block (RFBlock), which fuses re-weighted features to keep balance between them and high-level features from decoder. Additionally, a training strategy named self-transfer is used to effectively initialize encoders, which can stimulate different encoders to make full mining of modality-specific features. By the MMFNet, we combine multiple MRI to realize accurate segmentation of NPC. We implement extensive experiments and comparisons with the related methods to demonstrate its effectiveness and advantages. 

The main contributions of this paper can be summarized as followed:
\begin{itemize}
	\item[(1)] MMFNet, which is a multi-encoder single-decoder network, is proposed to capture local, global, cross-modality and modality-specific features from multiple modalities of MRI to realize accurate segmentation of NPC.
	\item[(2)] A novel fusion block, containing a 3D-CBAM and a RFBlock ,is designed in the MMFNet to learn complementary features and cross-modal interdependencies from multi-modality MRI. 3D-CBAM is specially designed for the recalibration of features from 3D medical images. RFBlock is proposed to fuse modality-specific features for further processing.
	\item[(3)] The training strategy named self-transfer can stimulate different encoders to make full mining of modality-specific features from multi-modality MRI.
\end{itemize}

The remaining paper is organized as followed. In section \ref{section:related_work}, related works will be reviewed. Section \ref{section:methodology} will introduce our proposed framework. Experimental results and analysis will be reported in section \ref{section:experiments and analysis}. Then in section \ref{section:discussion} we will set some ablation experiments to further discuss proposed method. Finally, a conclusion will be made in section \ref{section:conclusion}.

\section{Related work}
\label{section:related_work}
\subsection{Multi-modal fusion}
\label{section:related_work:multi-modal fusion}
In medical images, combining multi-modality images such as CT, T1 MRI, T2 MRI, etc. to realize multi-organ segmentation \cite{valindria2018multi} and lesion segmentation \cite{tseng2017joint,ma2018concatenated} is widely adopted due to distinct responses of different modalities datasets for different tissues. According to the review \cite{zhou2019review} of deep learning for medical image segmentation using multi-modality fusion, multi-modal segmentation network architectures can be categorized into input-level fusion network, layer-level fusion network and decision fusion network. The input-level fusion network \cite{havaei2017brain,kamnitsas2017efficient} stacks multi-modality images channel-wise and directly feeds them into neural network to make final decisions. In decision-level fusion segmentation networks, multiple pathways are set to process separately multi-modal images and the final features \cite{nie2016fully} or results \cite{kamnitsas2017ensembles} are combined for decision making. The layer-level fusion network \cite{valindria2018multi,tseng2017joint,dolz2018hyperdense,dolz2018ivd} fuses multi-source features in mediate layers to obtain complementary and interdependent features. Multi-encoder-based method \cite{valindria2018multi,tseng2017joint}, using multiple modality-specific encoders to extract features at various levels and feeding them into one single decoder, is the typical framework of layer-level fusion network. For the fusion of multi-source layers' features, besides directly merging encoders' features \cite{tseng2017joint}, linking features across multi-path \cite{dolz2018hyperdense} and linking features across multi-encoder \cite{dolz2018ivd} are also great strategies.

According to several researches \cite{valada2016deep,valindria2018multi}, methods based on multiple encoders have better capability to capture complementary and cross-modal interdependent features. Therefore, our proposed framework is based on multi-encoder-based method. Although multi-encoder-based methods can combine information from multi-modality datasets to capture complementary and interdependent features, some individual features of specific modality can still be ignored. Additionally, there is not a pre-trained powerful enough 3D network to extract general features from medical images. To address these problems, we present a training strategy named self-transfer to initialize encoders to make full mining of modality-specific features. And these features will be further fused to get interdependent cross-modality features. 

There also exist problems when we fuse multi-source features. The differences among low-level features from different modalities of MRI and the large imbalance in the channel numbers between low-level and high-level features can make network confused if we only simply merge them. Therefore, a fusion block is proposed to fuse low-level features and prepare fused ones for the fusion with high-level features. The fusion block can adaptively recalibrate low-level features from modality-specific encoders and fuse them into features with the same channel number of corresponding high-level features to keep balance between high-level and low-level features.

\subsection{Attention mechanism}
\label{section:related_work:attention mechanism}
In human perception, the information gained from different sensory channels will be weighted by attention mechanism, namely, greater weights will be ascribed to sensory streams providing reliable information from the world \cite{feldman2010attention}. Specially, in human visual attention mechanism, only a subset of sensory information will be selected by intermediate and higher visual processes to be further processed \cite{itti1998model}. The idea of attention mechanism has been successfully implemented in several DNN-based frameworks for image classification \cite{wang2017residual,hu2017squeeze,woo2018cbam}, image understanding \cite{chen2017sca}, target detection \cite{woo2018cbam}, etc.. 

For attention mechanism, channel attention modules and spatial attention modules are two major topics. Channel attention modules attempt to adaptively recalibrate channel-wise feature responses, while, spatial attention modules emphasize the usage of concentrating on ROIs. SENet \cite{hu2017squeeze} and Residual Attention Network \cite{wang2017residual} are respectively representatives of channel attention modules and spatial attention modules. In addition to using one of these two modules, there also exist methods to combine both of them. CBAM \cite{woo2018cbam} is a lightweight architecture simultaneously employs spatial and channel-wise attention to improve performance of DNN. In this method, both max-pooling outputs and average-pooling outputs are fed into a shared multi-layer perceptron (MLP) to obtain channel-wise attention. Meanwhile, similar pooling outputs along channel axis are fed into a convolutional block to produce spatial attention.

Inspired by attention mechanism, we propose 3D-CBAM, which is an attention framework specifically designed for 3D multi-modality medical images, to recalibrate multi-source features to reduce confusion when fusing them. It uses a channel attention block and a spatial attention block to highlight both informative features and ROIs. For the design of channel attention block, in order to provide more sufficient information for 3D medical images, besides max-values and average-values, standard deviations (stds) are also captured to make final channel-wise weights. Meanwhile, stds are also adopted to improve the performance of spatial channel block. Additionally, we respectively set three MLPs for these three kinds of features due to the huge differences among the distribution of average-values, max-values and stds.

\section{Methodology}
\label{section:methodology}
\begin{figure*}[]
	\centering	
	\includegraphics[width=1\textwidth]{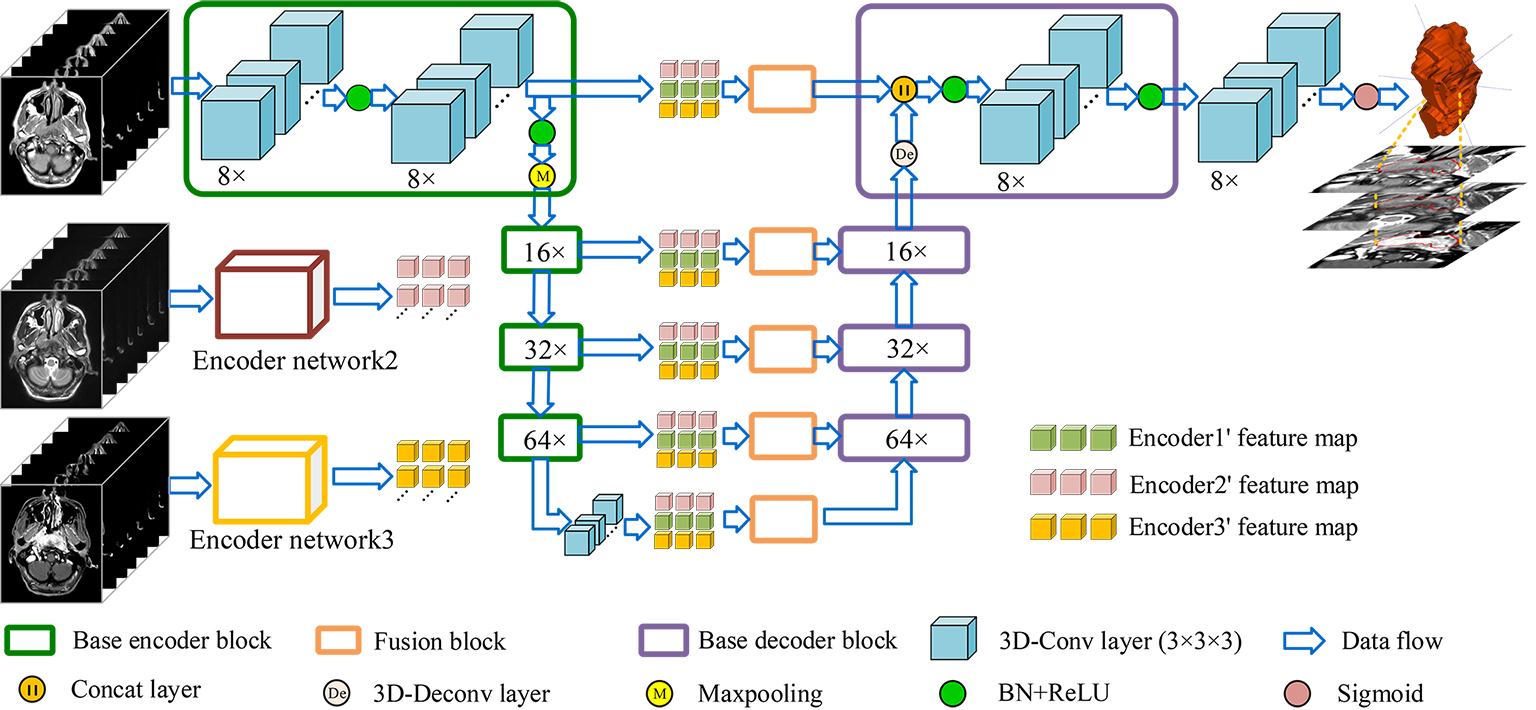}
	\caption{An illustration of our proposed framework.}
	\label{fig:framework}
\end{figure*}
As illustrated in Figure \ref{fig:framework}, our framework is an end-to-end fully convolutional network, containing three encoders to take 3D images from three modalities of MRI as inputs. The encoder network is a VGG-liked \cite{simonyan2014very} DNN, which stacks base block containing several 3D convolutional layers followed by max-pooling layers to get deeper features. And the decoder network uses 3D deconvolutional layers to upsample feature maps, the final output is a feature map with the same size as input. Both low-level features and high-level features, which are relevant to NPC segmentation, can be obtained by the design of multiple encoders and one single decoder. In order to effectively fuse low-level features from multi-modality MRI and keep balance between high-level and low-level features, a fusion block composed with 3D-CBAM and RFBlock is proposed to recalibrate and fuse multi-source low-level feature maps. For the training of network, we propose self-transfer to use pre-trained modality-specific encoders, which can capture individual modality-specific features from single modality MRI, as initial encoders of multi-modality model. The utilization of self-transfer can effectively improve the performance of encoders and make full mining of informative features from every modality of data.

\subsection{Base encoder-decoder network}
\label{section:methodology:base encoder-decoder network}
\begin{figure}[]
	\centering		
	\subfigure{
		\begin{minipage}[]{0.43\textwidth}
			\centering
			\includegraphics[width=\textwidth]{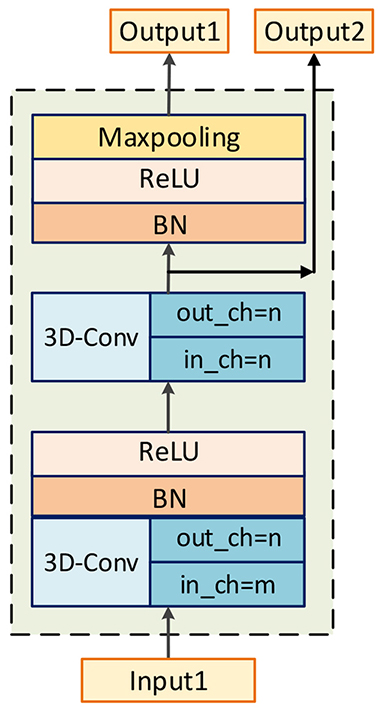}
		\end{minipage}
		\label{fig:encoder-decoder:encoder_block}
	}
	\subfigure{
		\begin{minipage}[]{0.45\textwidth}
			\centering
			\includegraphics[width=\textwidth]{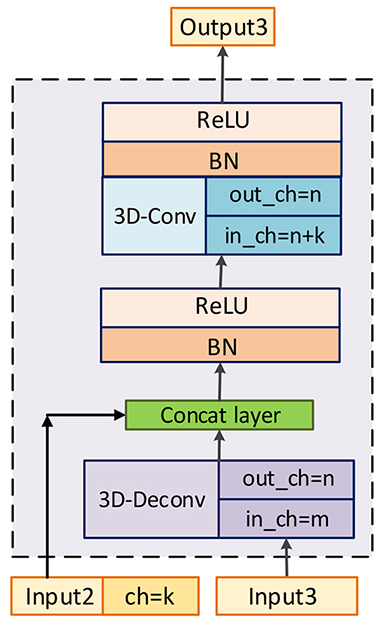}
		\end{minipage}
		\label{fig:encoder-decoder:decoder-block}
	}
	\caption{An illustration of base encoder block and base decoder block. Left: base encoder block. Right: base decoder block. Input1, Input3: features from previous block. Input2: features produced by corresponding encoder block. Output1, Output3: features, which will be fed into next block. Output2: features, which will be fed into the corresponding decoder block through skip connection layer.}
	\label{fig:encoder-decoder}
\end{figure}

Inspired by U-net \cite{ronneberger2015u}, our base encoder-decoder network can be seen as a U-net composed with 3D convolutional layers and 3D deconvolutional layers. Through stacking convolutional layers and max-pooling layers, encoder network can get larger receptive field, meanwhile, the spatial resolution becomes smaller. On the contrary, decoder network is composed with 3D deconvolutional layers to upsample feature maps, thus, the spatial resolution of features can recover to original scale when high-level features go through it. As described in \cite{gatys2016image}, higher layers capture high-level representations, which are necessary for recognize targets, and lower layers capture low-level representations such as texture, which play major roles in reducing the missing of tiny structures when we segment objects. Therefore, both content and style representations should be utilized to complete NPC segmentation. Thus, skip connection layers are adopted to combine low-level and high-level features. The architectures of base encoder block and decoder block are illustrated in Figure \ref{fig:encoder-decoder}.

\textbf{Base encoder network.} Our encoder network is a VGG-liked \cite{simonyan2014very} network, and the base block is composed with two 3D convolutional layers. According to \cite{szegedy2016rethinking}, the representation size should slightly decrease to avoid bottlenecks with enormous compression. Therefore, $3\times{3}\times{3}$ 3D convolutional layers and $2\times{2}\times{2}$ max-pooling layers are preferential choices to construct our network. And after convolution, a batch normalization layer and a ReLU layer are followed. There are two outputs produced by encoder block, one is for next encoder block, and the other one is for the corresponding decoder block to realize the combination of high-level and low-level features. There are totally four encoder blocks and the channel number of outputs ($Out\_ch$) is 8, 16, 32 and 64 respectively. It is worth mentioning that there is one single convolutional layer after final encoder block to refine features downsampled by encoder block, and the number of its channels is 64.

\textbf{Base decoder network.} The purpose of decoder network is to map high-level features to target modality. A 3D deconvolotional layer is utilized to upsample feature maps, then, a concatenation layer combines these features with low-level features from encoders with the assistance of skip connection layers. After merging, a convolutional layer is adopted to fuse these feature maps. The numbers of output's channels for decoder blocks are 64, 32, 16 and 8 respectively.

Finally, the final decoder block is followed by a convolutional layer with sigmoid as activation function to produce final segmentation results.

\textbf{Loss function.} Inspired by \cite{milletari2016v}, which presents Dice coefficient to effectively solve imbalance between the numbers of voxels of foreground and background, we apply Dice loss as network's optimization objective. We denote ground truth as $G$ and $P$ is denoted as predict results. The definition of Dice loss is shown as:
\begin{equation}
	Loss_{dice}=1-2\times{\frac{\sum_{i=1}^{N}{p_{i}g_{i}+\epsilon}}{\sum_{i=1}^{N}{p_{i}}+\sum_{i=1}^{N}{g_{i}}+\epsilon}}
\end{equation}
\begin{equation}
	p_{i}\in{P}
\end{equation}
\begin{equation}
	g_{i}\in{G}
\end{equation}
Where $\epsilon$ is smoothness term to avoid the risk of being divided by 0, and we set $\epsilon=1$ in our experiments.

\subsection{Fusion block}
\label{section:methodology:fusion block}
\begin{figure*}[]
	\centering		
	\subfigure[Channel attention block]{
		\begin{minipage}[]{0.49\textwidth}
			\centering
			\includegraphics[width=\textwidth]{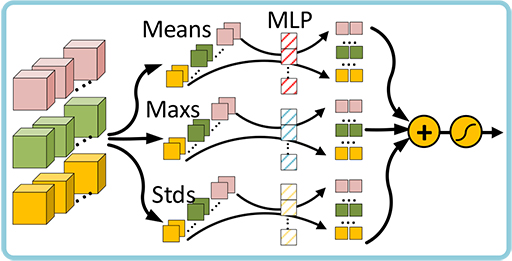}
		\end{minipage}
		\label{fig:fusion_block:channel_attention}
	}
	\subfigure[Spatial attention block]{
		\begin{minipage}[]{0.40\textwidth}
			\centering
			\includegraphics[width=\textwidth]{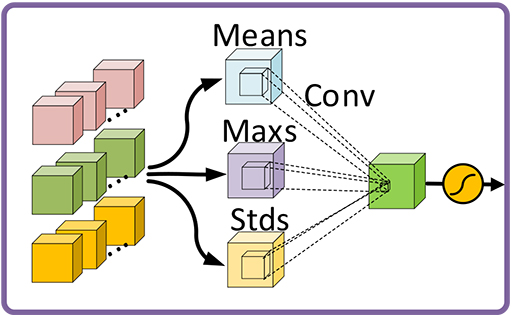}
		\end{minipage}
		\label{fig:fusion_block:spatial_attention}
	}
	\subfigure[Fusion block]{
		\begin{minipage}[]{0.97\textwidth}
			\centering
			\includegraphics[width=\textwidth]{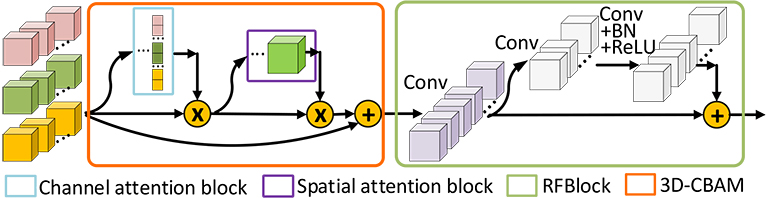}
		\end{minipage}
		\label{fig:fusion_block:fusion_block}
	}
	\caption{The architecture of fusion block. }
	\label{fig:fusion_block}
\end{figure*}

The purpose of fusion block is to effectively recalibrate and fuse low-level features from different modalities of MRI before merging them with high-level features. It is a hard task to directly fuse low-level features from multiple MRI, which vary greatly from each other due to the varied responses to different tissues of multi-modality MRI. Therefore, the fusion block will firstly re-weighting features and highlight regions that are greatly relevant to NPC with the assistance of 3D-CBAM. 3D-CBAM is composed with a channel attention block, which focuses on 'what' are meaningful features, and a spatial attention block, which focuses on ‘where’ is an interesting part. After recalibrating low-level features, an RFBlock is utilized to fuse them into ones with the same channel number of corresponding high-level features to keep balance between them. The architectures of them are shown in Figure \ref{fig:fusion_block}.

Given three intermediate feature maps from multiple encoders $F_{1}, F_{2}, F_{3} \in{R^{C\times{D}\times{H}\times{W}}}$ as inputs. We firstly merge them on channel axis to obtain total original feature maps $F_{ori}\in{R^{3C\times{D}\times{H}\times{W}}}$. And we denote the final fused features as $F_{fused}$, the overall fusing process can be summarized as followed:
\begin{equation}
	F_{c}=W_{c}(F_{ori})\otimes{F_{ori}}
\end{equation}
\begin{equation}
	F_{s}=W_{s}(F_{c})\otimes{F_{c}}
\end{equation}
\begin{equation}
	F_{fused}=Res(F_{s}+F_{ori})
\end{equation}
We denote $\otimes$ as element-wise multiplication, which will automatically broadcast spatial attention weights ($W_{s}$) and channel attention weights ($W_{c}$) to fit input feature maps. Meanwhile, $F_{c}$ and $F_{s}$ are denoted as feature maps after being refined by channel attention block and spatial attention block respectively. $F_{fused}$ is the final output, which is fused and refined by residual fusion block.

\textbf{Channel attention block.} The focus of channel attention block is to selectively emphasize feature maps, which are meaningful for final predictions. SENet \cite{hu2017squeeze} utilizes global average pooling to capture average-values of each feature map, and feeds them into a MLP to get weights for every channel. Compared to SENet, the channel attention block of CBAM \cite{woo2018cbam} uses global max-pooling layers to get max-values of feature maps in additional to capturing average-values using global average-pooling layers. And both average-values and max-values are fed into a shared MLP. Then, the output vectors are combined by add operation. 

In this paper, in order to better represent the global features for 3D medical images, we capture stds of every three-dimension features and combine them with average-values and max-values to produce weights for every channel feature. We denote obtained stds, average-values and max-values as $F_{std}, F_{avg}, F_{max} \in{R^{3c\times{1}\times{1}\times{1}}}$. In terms of MLP, because there exist great gaps among the distribution of stds, average-values and max-values, we respectively set three MLPs for them. All of these MLPs are composed with one hidden layer setting hidden activation size as $R^{3c/r\times{1}\times{1}\times{1}}$, where $r$ is the reduction ratio and we set $r=12$ in our experiments. And the final output channel weights $W_{c}$ has $3c$ values for each channel feature. The formulation for $W_{c}$ is shown as followed:
\begin{equation}
	W_{c}(F_{ori})=\sigma{(MLP_{avg}(F_{avg})+MLP_{max}(F_{max})+MLP_{std}(F_{std}))}
\end{equation}
\begin{equation}
	F_{avg}=AvgPool(F_{ori})
\end{equation}
\begin{equation}
	F_{max}=MaxPool(F_{ori})
\end{equation}
\begin{equation}
	F_{std}=StdPool(F_{ori})
	=(AvgPool((F-AvgPool(F_{ori}))^{2}))^{1/2}
\end{equation}
Where $\sigma$ is the sigmoid activation to produce channel-wise weights ranged from 0 to 1. Figure \ref{fig:fusion_block:channel_attention} shows the architecture of channel attention block.

\textbf{Spatial attention block.} The purpose of spatial attention block is to utilize feature maps after channel-wise refining to obtain 3D spatial attention map ($W_{s}$). On the basis of previous work of CBAM, we capture stds, average-values and max-values along the channel axis and concatenate them to generate three 3D feature blocks. And these features are fed into a $3\times{3}\times{3}$ 3D convolutional layer with sigmoid as activation to produce $W_{s}$. Through element-wise multiplication, informative regions will be effectively highlighted. The architecture of spatial attention block is shown in Figure \ref{fig:fusion_block:spatial_attention}, and the process can be summarized as followed:
\begin{equation}
	W_{s}=\sigma{(f^{3\times{3}\times{3}}([AvgPool(F_{c});MaxPool(F_{c});StdPool(F_{c})]))}
\end{equation}
Where $f^{3\times{3}\times{3}}$ is denoted as one single convolutional layer with kernel size of $3\times{3}\times{3}$ and the number of output's channels is 1.

\textbf{Residual fusion block.} After highlighting informative features and ROIs, a residual fusion block is constructed to fuse and refine low-level features. It is worth mentioning that the numbers of channels are $3c$ for $(F_{ori}+F_{s})$, while the corresponding high-level features, which is prepared to combine with fused features, only has $c$ channels. Therefore, in order to keep balance between low-level and high-level features, an $1\times{1}\times{1}$ convolutional layer with $c$ channel outputs is utilized to fuse feature maps and reduce channel number firstly. Then, a residual block \cite{he2016deep} is adopted to refine feature maps. This block is composed with two convolutional layers, both of them have $1\times{3}\times{3}$ kernels and the first one is followed by a batch normalization layer and a ReLU layer. We can summary this process using following equations:
\begin{equation}
	F_{r}=f^{1\times{1}\times{1}}(F_{ori}+F_{s})
\end{equation}
\begin{equation}
	F_{fused}=F_{r}+f^{1\times{3}\times{3}}(f^{1\times{3}\times{3}*}(F_{r}))
\end{equation}
Where $F_{r}$ is feature maps after $1\times{1}\times{1}$ convolutional layer, and $f^{1\times{3}\times{3}*}$ is denoted as $1\times{3}\times{3}$ convolutional layer followed by a batch normalization layer and a ReLU layer.

\subsection{Self-transfer learning}
\label{section:methodology:self-transfer learning}
\begin{figure}[]
	\centering	
	\includegraphics[width=0.95\textwidth]{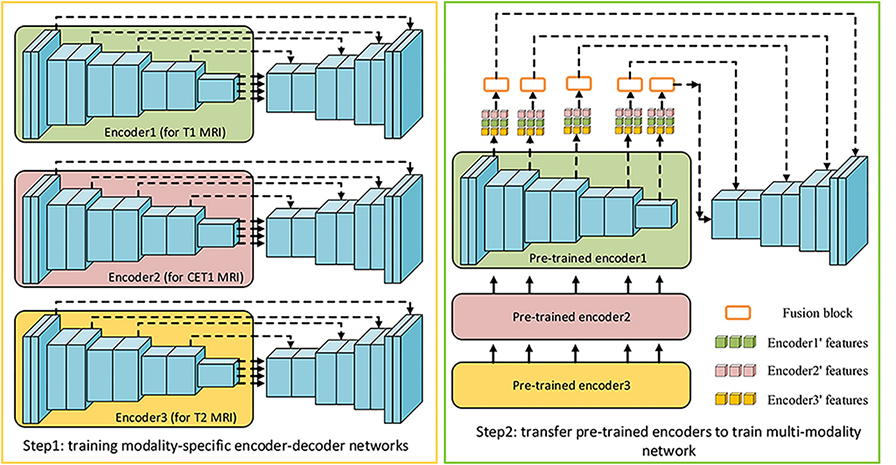}
	\caption{An illustration of the self-transfer learning.}
	\label{fig:self_transfer}
\end{figure}

Transfer learning \cite{yosinski2014transferable}, utilizing a powerful pre-trained network as features' extractor, is a popular trick to improve performance of new systems. Hence, using a network pre-trained in imagenet as an encoder for a segmentation network is a common operation in natural images \cite{yang2018denseaspp,zhao2017pyramid}. However, there is not a powerful enough 3D pre-trained model can be set as the initial features' extractor for various 3D medical images due to their complexity and various imaging technologies. Especially for multi-modality MRI, images of each modality have their own specific imaging styles, it's hard to obtain a features' extractor, which can be generalized to all of them. Additionally, by the design of multi-encoder single-decoder network, complementary information and cross-modal interdependencies can be extracted, while some individual features of specific modality may be ignored. To address these problems, we propose an initialization trick named self-transfer to effectively initialize encoders and make full mining of features of different modalities of MRI. According to experimental results, performance of multi-encoder-based models can obtain considerable improvements by using self-transfer.

Specifically speaking, a modality-specific model can effectively capture individual informative features from one single modality of data, while a multi-modality model aims to obtain interdependent and complementary information from multiple-modality datasets. As a result, some individual features of one single modality may be ignored in multi-modality model. Therefore, we propose self-transfer to fully mine modality-specific features. Figure \ref{fig:self_transfer} is the illustration of self-transfer. The first step is to respectively train three modality-specific encoder-decoder models. Then, these pre-trained encoders will be used as the initial encoders for multi-modality model. Compared to original encoders with random initialization, these encoders have greater power to make full mining of individual features from specific modality of MRI. Meanwhile, the fusion block and decoder can effectively fuse these features to obtain informative features for final predictions. We will set several experiments to demonstrate that self-transfer can enhance the segmentation systems in the following paper.

\section{Experiments and Comparations}
\label{section:experiments and analysis}
\subsection{Dataset and Preprocessing}
\label{section:experiments and analysis:dataset and analysis}
\textbf{Dataset.} Three modalities of MRI of T1, CET1 and T2 of 149 patients are acquired at Shandong Cancer Hospital Affiliated to Shandong University. These patients are scanned with Philips Ingenia 3.0T MRI system. Both CET1 and T2 are aligned to T1. Different 3D images might have different resolutions, for example, they might have sampling spacings from $0.33mm$ and $0.69 mm$ along X and Y axes, and $3.5mm$ to $5.5mm$ along Z axis (i. e, the spacing between adjacent sectional images). Thus, we resample these 3D images such that they all have the same spacing along X and Y axes ($0.5 mm$) while keeping their respective Z-spacing unchanged to avoid obtaining unreal images. After resampling, the sizes of 3D images are ranged from $[24,387,387]$ to $[56,520,520]$. The ground truth is created by an experienced radiologist and marked slice by slice. 

\begin{figure}[]
	\centering	
	\includegraphics[width=0.95\textwidth]{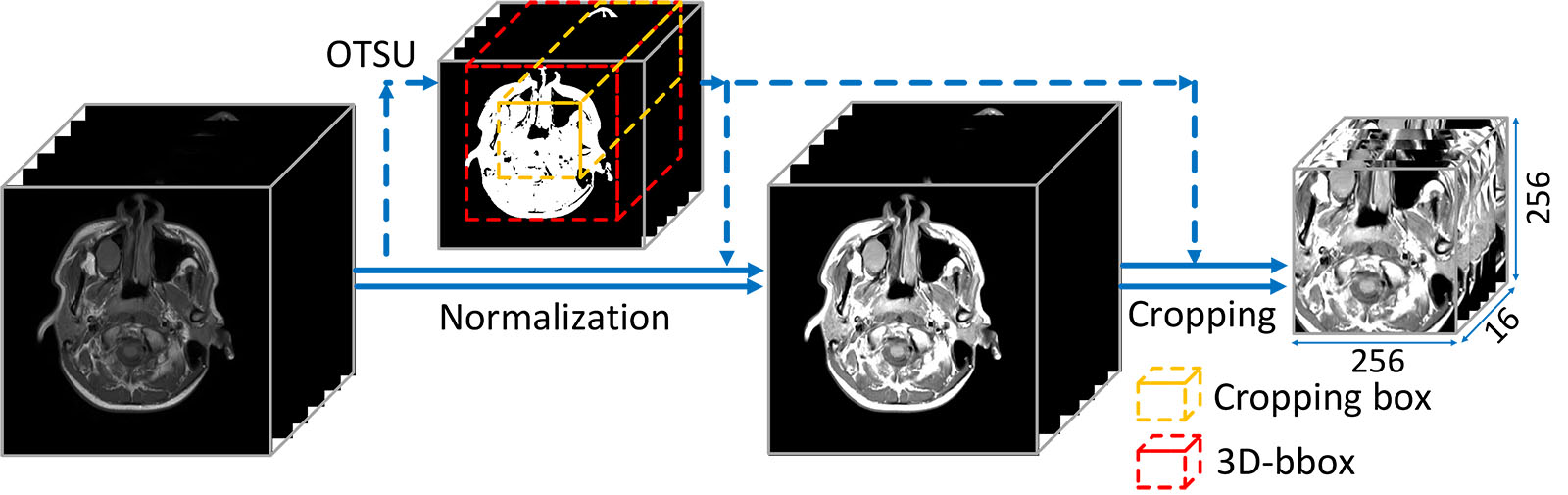}
	\caption{An illustration of preprocessing.}
	\label{fig:preprocessing}
\end{figure}

\textbf{Preprocessing.} The preprocessing mainly contains intensity normalization and ROIs cropping. We utilize the intra-body intensity normalization proposed in \cite{huang20183d}, which effectively deals with the differences caused by imaging configurations and the influences of inconsistent body-to-background ratios. After normalization, according to the distribution histogram of normalized data, we clip (limit) values to reduce the complexity of data. Through statistical analysis for values in NPC regions, we set $[-2,1]$, $[-3,2]$ and $[-2,3.5]$ for T1, T2 and CET1 respectively.

We use a sliding window to crop ROIs and feed them into network. Specifically speaking, we firstly use OTSU to get the 3D bounding box (3D-bbox) of original MRI and then we crop a $16\times{256}\times{256}$ region according to the central point of 3D-bbox. The sliding window slides on the Z axis and the sliding-step is 4. When feeding training samples, we perform on-the-fly data augmentation. Operations including random central offset from $(-4,-32,-32)$ to $(4,32,32)$, random vertical flipping and random rotation on the XY plane from $-5^{\circ}$ to $5^{\circ}$ are applied. The illustration of preprocessing is shown in Figure \ref{fig:preprocessing}.

\subsection{Evaluation metrics}
\label{section:experiments and analysis: evaluation metrics}
\emph{1) Dice Similarity Coefficient (DSC):} The dice similarity coefficient is designed to evaluate the overlap rate of predict results and ground truth. DSC can be written as: 
\begin{equation}
	DSC(P,G)=\frac{2\times{|P\cap{G}|}}{|P|+|G|}
\end{equation}
DSC ranges from 0 to 1, and the better predict result will have a larger DSC.

\emph{2) Average symmetric surface distance (ASD):} It is a measure to show average of all distance between two image volumes. The $ASD$ is denoted as below:
\begin{equation}
ASD=\frac{1}{|B_P|+|B_G|}\times{(\sum_{x\in{B_P}}d(x,B_G)+\sum_{y\in{B_GT}}d(y,B_P))}
\end{equation}
\begin{equation}
	d(x,A)=\min\limits_{y\in{A}}||x,y||
\end{equation}
Where $||.||$ denotes Euclidean distance. $B_P$ and $B_G$ denote the volume's surface of predictions ($P$) and ground truths ($G$). And $|.|$ is the number of points.

\emph{3) Hausdorff Distance (HD):} It shows the greatest value of distances from a point in one volume's surface to the closest point in the other surface. The formulation is presented as followed:
\begin{equation}
	H(B_P,B_G)=max(h(B_P,B_G),h(B_G,B_P))
\end{equation}
\begin{equation}
	h(B_P,B_G)=\max\limits_{x\in{B_P}}(\min\limits_{y\in{B_G}}(||x,y||))
\end{equation}
\begin{equation}
h(B_G,B_P)=\max\limits_{x\in{B_G}}(\min\limits_{y\in{B_P}}(||x,y||))
\end{equation}

\subsection{Experiments setting}
\label{section:experiments and analysis:Experiments setting}
The proposed MMFNet (MMFNet + multi-MLP + stdPool + self-transfer) is composed with three encoders for three modalities of MRI (T1, T2 and CET1), fusion blocks and one single decoder. There are three MLPs in channel attention block. And both channel attention block and spatial attention block contain std-pooling, max-pooling and average-pooling. In the training stage, we firstly train three modality-specific networks and then transfer the pre-trained encoders as the initial encoders for MMFNet. For the training of these modality-specific networks, we set Adam \cite{kingma2014adam} as optimizer at a learning rate of $10^{-3}$. While for the training of MMFNet, we firstly freeze encoders' parameters in the first five epochs to warm up decoder with Adam at a learning rate of $10^{-3}$. After decoder has been warmed up, we update both decoder and encoders of MMFNet with learning rate of $10^{-4}$. The batch size for all MMFNet and modality-specific networks is 8. We set max epochs as 100 and networks will be updated 75 times each epoch.

MMFNet is evaluated in five-fold cross validation. And 25\% of training data will be divided as validation data to choose best model and alleviate overfitting. Meanwhile, we use early-stopping strategy to stop training if validation loss does not decrease over 10 epochs to reduce overfitting.

Our experiments are performed on a workstation platform with Intel(R) Xeon(R) CPU E5-2620 v4 @ 2.10GHz, 64GB RAM and 2x NVIDIA Titan Xp GPU with 12GB GPU memory. The code is implemented with pytorch 0.4.1 in Windows 10.

\subsection{Comparative experiments}
\label{section:experiments and analysis:Comparative experiments}
In this subsection, we set extensive comparative experiments to show the performance of MMFNet. These methods are described as followed:

1) Patch-based CNN \cite{wang2018automatic,ma2018automated}, which utilizes a sliding window to capture patches of single-modality MRI to make decision whether the center point belongs to tumor. 2) Multi-modality patch-based CNN \cite{ma2018discriminative} is a great method integrating multiple patch-based CNN into a Siamese-like sub-network to make predictions based on multi-modality medical images of NPC. 3) The U-net \cite{ronneberger2015u}, which is a novel end-to-end network for segmentation and is widely used in segmentation of medical images. 4) The 3D U-net \cite{cciccek20163d}, which is the 3D vision of U-net. 5) Input-level fusion network \cite{havaei2017brain}, which stacks different modalities of MRI channel-wise as input for network. We concentrate three volumes of MRI as different input channel, the architectures of encoder and decoder are as described in section \ref{section:methodology:base encoder-decoder network}, while we set the channel number of encoder as three times of original encoder to keep the number of low-level features is the same as MMFNet. 6) Merging encoders' features \cite{tseng2017joint}, setting three modality-specific encoders to capture low-level features and a decoder to fuse low-level and high-level features. 7) Linking features across multi-path \cite{dolz2018hyperdense}, which builds multiple streams for different modalities of MRI and links features across these streams. 8) Linking features across multi-encoder, setting individual encoders for every modality MRI and skip connections are built across different encoders. 9) Decision-level fusion, fusing final features from modality-specific paths to make final decisions \cite{nie2016fully}.

\begin{figure*}[]
	\centering	
	\subfigure[DSC=0.915]{
		\begin{minipage}{0.24\textwidth}
			\centering
			\includegraphics[width=\textwidth]{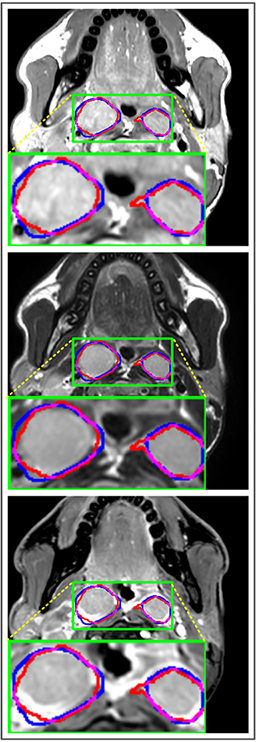}
		\end{minipage}
		\label{fig:2D_slice_for_MMFNet:088_29_09205}
	}	
	\hspace{-0.4cm}
	\subfigure[DSC=0.870]{
		\begin{minipage}{0.24\textwidth}
			\centering
			\includegraphics[width=\textwidth]{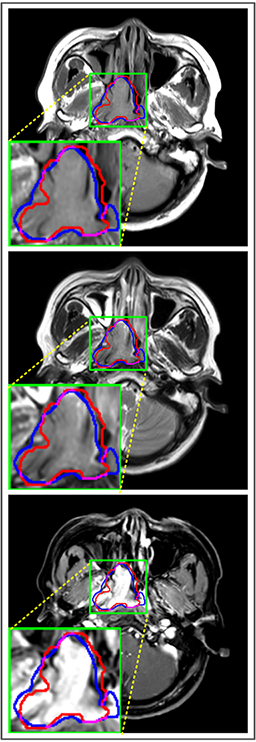}
		\end{minipage}
		\label{fig:2D_slice_for_MMFNet:155_37_08704}
	}	
	\hspace{-0.4cm}
	\subfigure[DSC=0.902]{
		\begin{minipage}{0.24\textwidth}
			\centering
			\includegraphics[width=\textwidth]{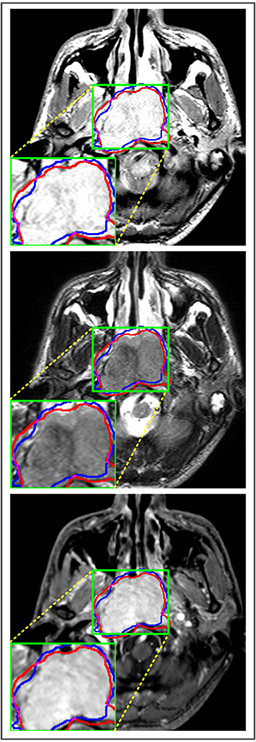}
		\end{minipage}
		\label{fig:2D_slice_for_MMFNet:159_13_09028}
	}	
	\hspace{-0.4cm}
	\subfigure[DSC=0.898]{
		\begin{minipage}{0.24\textwidth}
			\centering
			\includegraphics[width=\textwidth]{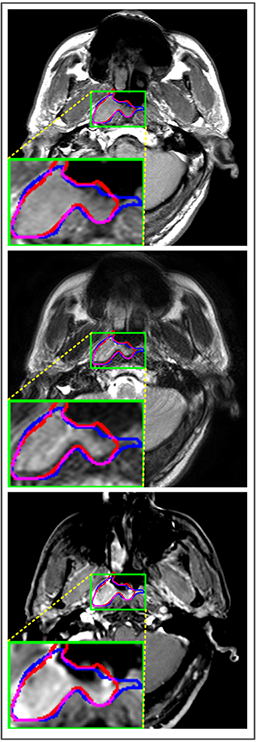}
		\end{minipage}
		\label{fig:2D_slice_for_MMFNet:170_33_08985}
	}	
	
	\caption{Predicted results of proposed MMFNet in 2D images. There are corresponding T1, T2 and CET1 images from top to down. Boundaries created by radiologists are marked in red line, and the predicted boundaries are shown in blue line. The $DSC$ value is the dice similarity coefficient of this single slice.}
	\label{fig:2D_slice_for_MMFNet}
\end{figure*}

\begin{figure*}[]
	\centering	
	\subfigure[DSC=0.860]{
		\begin{minipage}{0.24\textwidth}
			\centering
			\includegraphics[width=\textwidth]{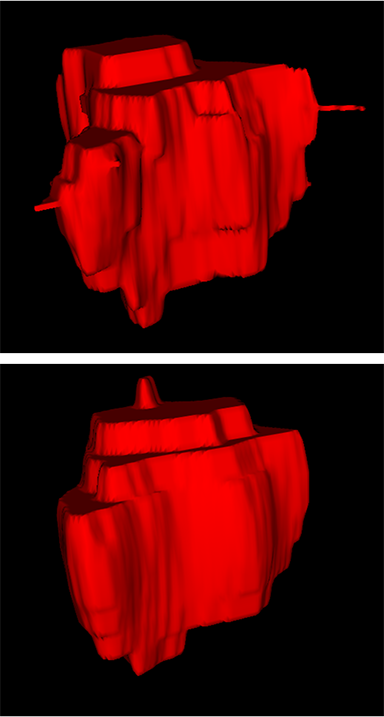}
		\end{minipage}
		\label{fig:3D_result_for_MMFNet:111}
	}	
	\hspace{-0.4cm}
	\subfigure[DSC=0.853]{
		\begin{minipage}{0.24\textwidth}
			\centering
			\includegraphics[width=\textwidth]{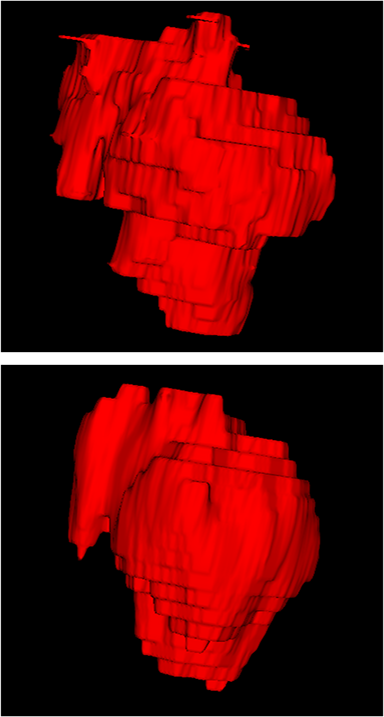}
		\end{minipage}
		\label{fig:3D_result_for_MMFNet:164}
	}	
	\hspace{-0.4cm}
	\subfigure[DSC=0.821]{
		\begin{minipage}{0.24\textwidth}
			\centering
			\includegraphics[width=\textwidth]{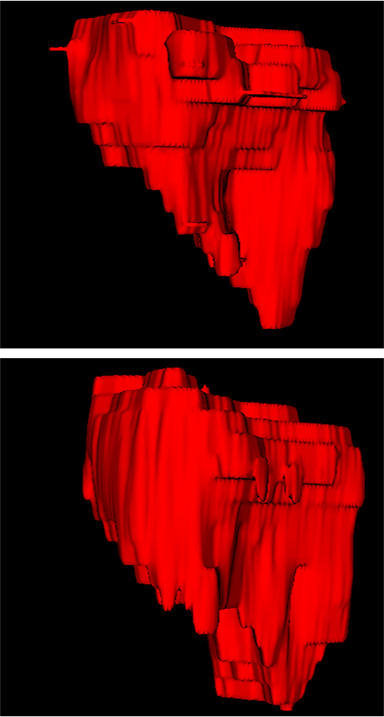}
		\end{minipage}
		\label{fig:3D_result_for_MMFNet:159}
	}	
	\hspace{-0.4cm}
	\subfigure[DSC=0.847]{
		\begin{minipage}{0.24\textwidth}
			\centering
			\includegraphics[width=\textwidth]{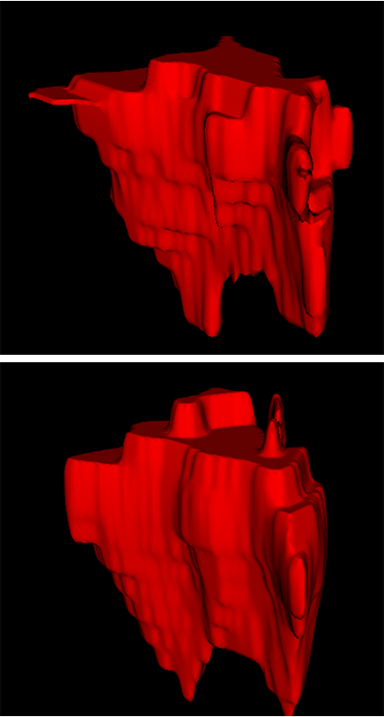}
		\end{minipage}
		\label{fig:3D_result_for_MMFNet:157}
	}	
	
	\caption{Examples of 3D predicted results for MMFNet. The first row are ground truth, and the second row are masks predicted by MMFNet.}
	\label{fig:3D_result_for_MMFNet}
\end{figure*}

\begin{figure*}[]
	\centering
	\vspace{-1cm}	
	\subfigure[]{
		\begin{minipage}{0.24\textwidth}
			\centering
			\includegraphics[width=\textwidth]{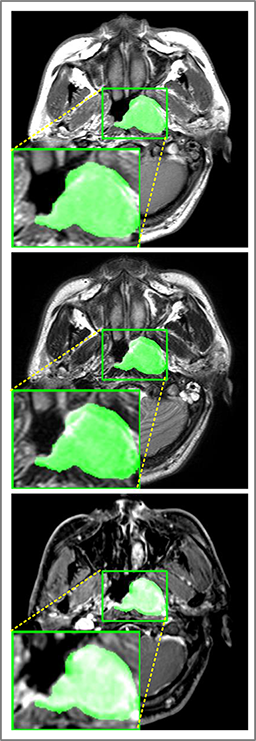}
		\end{minipage}
		\label{fig:2D result:ground truth}
	}	
	\hspace{-0.4cm}
	\subfigure[]{
		\begin{minipage}{0.24\textwidth}
			\centering
			\includegraphics[width=\textwidth]{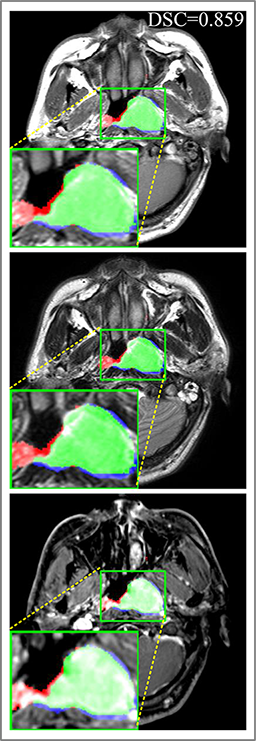}
		\end{minipage}
		\label{fig:2D result:patch-based_CET1}
	}
	\hspace{-0.4cm}
	\subfigure[]{
		\begin{minipage}{0.24\textwidth}
			\centering
			\includegraphics[width=\textwidth]{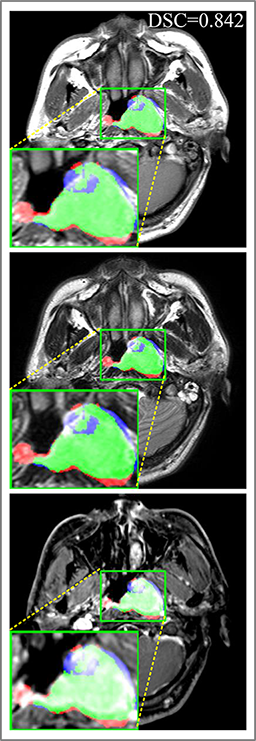}
		\end{minipage}
		\label{fig:2D result:multi-modality patch-based}
	}
	\hspace{-0.4cm}
	\subfigure[]{
		\begin{minipage}{0.24\textwidth}
			\centering
			\includegraphics[width=\textwidth]{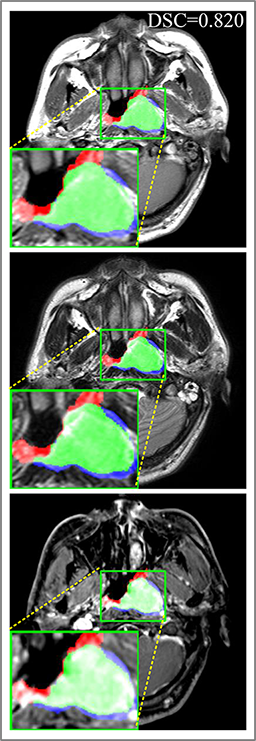}
		\end{minipage}
		\label{fig:2D result:Unet_CET1}
	}
	\hspace{-0.4cm}
	\subfigure[]{
		\begin{minipage}{0.24\textwidth}
			\centering
			\includegraphics[width=\textwidth]{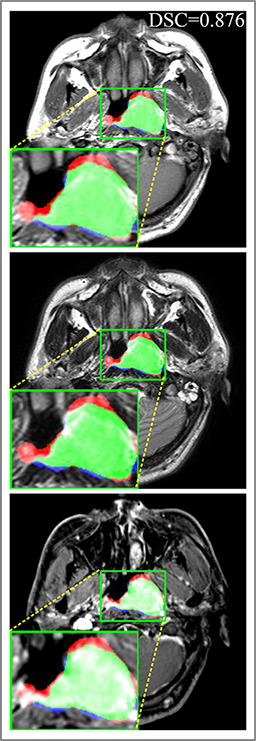}
		\end{minipage}
		\label{fig:2D result:modality-specific_CET1}
	}
	\hspace{-0.4cm}
	\subfigure[]{
		\begin{minipage}{0.24\textwidth}
			\centering
			\includegraphics[width=\textwidth]{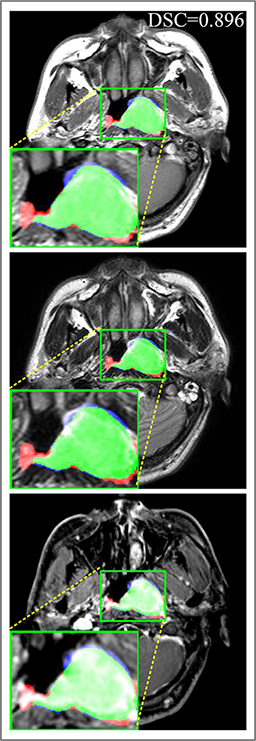}
		\end{minipage}
		\label{fig:2D result:multi-modality single-encoder}
	}	
	\hspace{-0.4cm}
	\subfigure[]{
		\begin{minipage}{0.24\textwidth}
			\centering
			\includegraphics[width=\textwidth]{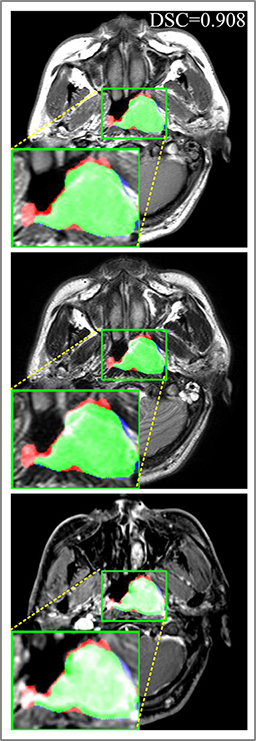}
		\end{minipage}
		\label{fig:2D result:multi-modality-multi-encoder}
	}
	\hspace{-0.4cm}
	\subfigure[]{
		\begin{minipage}{0.24\textwidth}
			\centering
			\includegraphics[width=\textwidth]{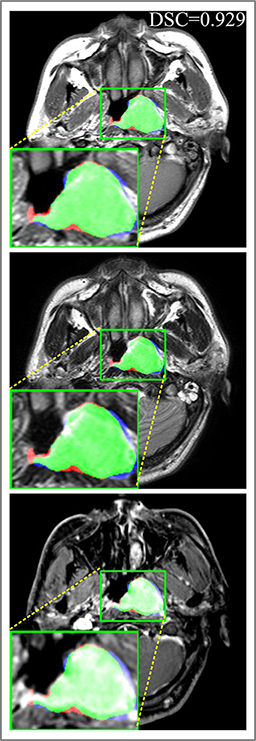}
		\end{minipage}
		\label{fig:2D result:MMFNet_final}
	}
	
	\caption{Predicted results in one single slice image, there are corresponding T1, T2 and CET1 images from top to down. Green regions denote $TP$ points, red and blue regions denote $FP$ and $FN$ point. (a) Ground truth. (b) Patch-based CNN (CET1). (c) Multi-modality patch-based CNN. (d) U-net (CET1). (e) 3D U-net (CET1). (f) Input-level fusion. (g) Merging encoders' features. (h) MMFNet + multi-MLP + stdPool + self-transfer.}
	\label{fig:2D result}
\end{figure*}

\begin{figure*}[]
	\centering	
	\subfigure[]{
		\begin{minipage}{0.24\textwidth}
			\centering
			\includegraphics[width=\textwidth]{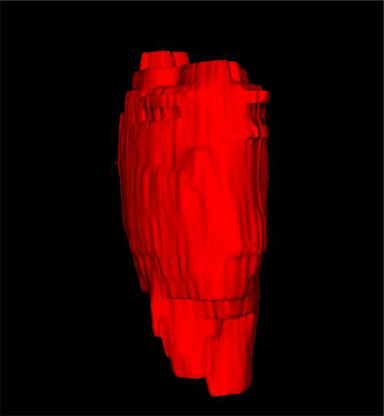}
		\end{minipage}
		\label{fig:3D result:ground truth}
	}	
	\hspace{-0.4cm}
	\subfigure[]{
		\begin{minipage}{0.24\textwidth}
			\centering
			\includegraphics[width=\textwidth]{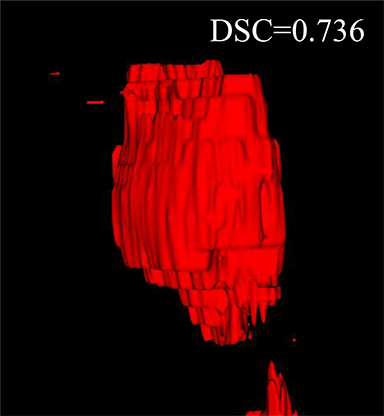}
		\end{minipage}
		\label{fig:3D result:patch-based_CET1}
	}
	\hspace{-0.4cm}
	\subfigure[]{
		\begin{minipage}{0.24\textwidth}
			\centering
			\includegraphics[width=\textwidth]{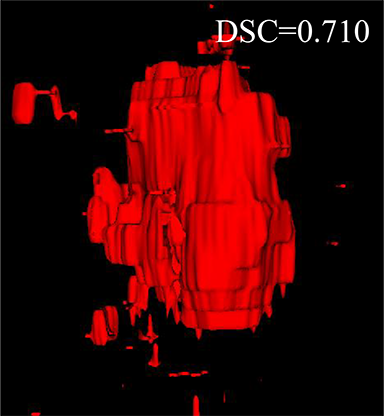}
		\end{minipage}
		\label{fig:3D result:multi-modality-patch-based}
	}
	\hspace{-0.4cm}
	\subfigure[]{
		\begin{minipage}{0.24\textwidth}
			\centering
			\includegraphics[width=\textwidth]{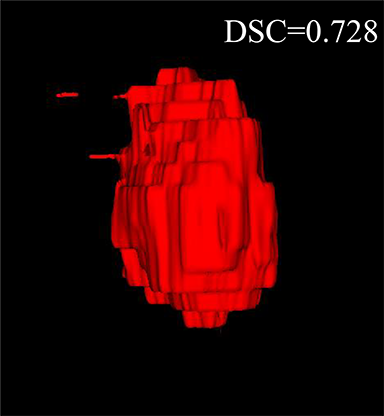}
		\end{minipage}
		\label{fig:3D result:Unet_CET1}
	}
	\hspace{-0.4cm}
	\subfigure[]{
		\begin{minipage}{0.24\textwidth}
			\centering
			\includegraphics[width=\textwidth]{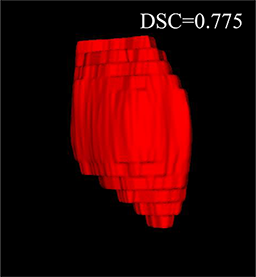}
		\end{minipage}
		\label{fig:3D result:3D-Unet_CET1}
	}
	\hspace{-0.4cm}
	\subfigure[]{
		\begin{minipage}{0.24\textwidth}
			\centering
			\includegraphics[width=\textwidth]{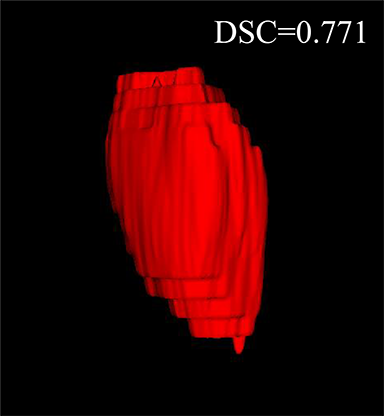}
		\end{minipage}
		\label{fig:3D result:single-encoder-multi-modality}
	}
	\hspace{-0.4cm}
	\subfigure[]{
		\begin{minipage}{0.24\textwidth}
			\centering
			\includegraphics[width=\textwidth]{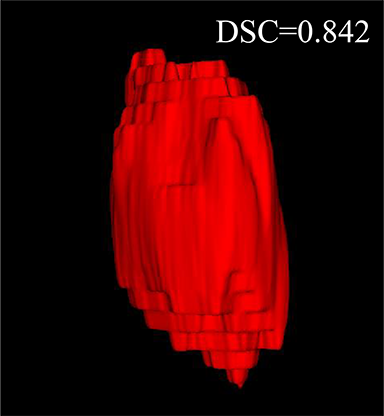}
		\end{minipage}
		\label{fig:3D result:multi-encoder-multi-modality}
	}
	\hspace{-0.4cm}
	\subfigure[]{
		\begin{minipage}{0.24\textwidth}
			\centering
			\includegraphics[width=\textwidth]{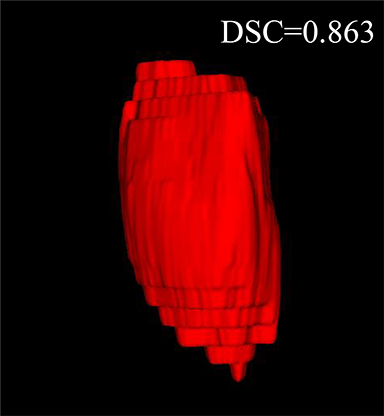}
		\end{minipage}
		\label{fig:3D result:MMFNet}
	}
	\caption{Examples of 3D predicted results for different methods. (a) Ground truth. (b) Patch-based CNN (CET1). (c) Multi-modality patch-based CNN. (d) U-net (CET1). (e) 3D U-net (CET1). (f) Input-level fusion. (g) Merging encoders' features. (h) MMFNet + multi-MLP + stdPool + self-transfer.}
	\label{fig:3D result}
\end{figure*}

\begin{table}
	\footnotesize
	\renewcommand\tabcolsep{4.0pt}
	\caption{Comparison of NPC segmentation results using different methods.}
	\centering
	\label{table:evaluation metrics}
	\begin{tabular}{lccccc}
		\toprule[1.5pt]
		Method & $mean DSC(\%)$ & $mean ASD(mm)$ & $mean HD(mm)$\\
		\midrule
		\midrule
		Patch-based CNN (T1) &$52.66\pm13.70$  &$17.66\pm9.43$   &$109.32\pm30.61$\\
		Patch-based CNN (T2) &$48.75\pm15.14$  &$21.86\pm14.76$  &$106.79\pm33.93$\\
		Patch-based CNN (CET1) &$56.04\pm13.64$&$8.64\pm6.74$    &$64.76\pm28.22$\\
		Multi-modality patch-based CNN &$60.12\pm15.03$ &$10.59\pm8.34$ &$99.49\pm30.33$\\
		\midrule
		U-net (T1) &$59.06\pm14.17$ &$14.60\pm9.93$ &$102.81\pm29.63$\\
		U-net (T2) &$55.48\pm14.67$ &$13.35\pm10.41$ &$98.20\pm34.67$\\
		U-net (CET1) &$59.39\pm13.53$&$6.10\pm6.41$ &$54.36\pm29.23$\\
		\midrule
		3D U-net (T1) & $67.19\pm14.84$ & $2.93\pm3.85$ & $19.98\pm14.87$ \\
		3D U-net (T2) & $64.68\pm15.42$ & $3.51\pm6.74$ & $21.76\pm20.48$ \\
		3D U-net (CET1) & $61.11\pm16.90$ & $3.23\pm2.75$ & $21.34\pm10.64$ \\
		\midrule
		Input-level fusion & $64.89\pm14.21$ & $4.22\pm4.51$ & $24.56\pm17.22$ \\
		Merging encoders' features & $69.74\pm10.95$ & $3.17\pm3.17$ & $30.18\pm32.89$ \\
		Linking features across multi-path & $69.06\pm12.52$ & $3.01\pm5.01$ & $19.74\pm12.94$ \\
		Linking features across multi-encoder & $70.63\pm10.11$ & $2.85\pm4.02$ & $19.70\pm13.28$ \\
		Decision-level fusion & $69.65\pm12.24$ & $2.84\pm1.91$ & $23.55\pm14.32$ \\
		MMFNet+multi-MLP+stdPool+self-transfer& $\textbf{72.38}\pm{10.99}$ & $\textbf{2.07}\pm2.32$ & $\textbf{18.31}\pm16.73$ \\
		\bottomrule[1.5pt]
	\end{tabular}
\end{table}

\subsection{Results}
\label{section:experiments and analysis:Results}

\textbf{Comparison with ground truth.} Some predicted results of MMFNet are shown in 2D images and 3D images in Figure \ref{fig:2D_slice_for_MMFNet} and Figure \ref{fig:3D_result_for_MMFNet}. As shown in these figures, although the shape and size of NPC are varied from each other, MMFNet can still accurately determine the regions of NPC and obtain the accurate contours of tumors. Through analyzing 2D images in figure \ref{fig:2D_slice_for_MMFNet}, MMFNet has a capacity to fuse multi-modality MRI to reduce the confusion brought by intensity' similarity between nearby tissues and NPC. The values of $meanDSC$, $meanASD$ and $meanHD$ of MMFNet are shown in Table \ref{table:evaluation metrics}. MMFNet can reach the best results with $DSC=72.38\%, meanASD=2.07mm$, and $meanHD=18.31mm$.

\textbf{Comparison with related works.} Table \ref{table:evaluation metrics} reports the values of $meanDSC$, $meanASD$ and $meanHD$ for different methods. Predicted masks of different methods are illustrated in Figure \ref{fig:2D result} and Figure \ref{fig:3D result}, which respectively present results in 2D and 3D images. Through comprehensively analyzing these results, the proposed MMFNet actually have the following properties:

\textbf{(i)} It directly fuses 3D MRI images rather than 2D slices. Thus, it can effectively use meaningful information from neighboring slices of MRI to realize NPC segmentation. As shown in Table \ref{table:ablation experiments}, MMFNet can bring $12.26\%$, $8.52mm$ and $81.19mm$ improvements in $meanDSC, meanASD$ and $meanHD$ compared to the best method based on 2D images (Multi-modality patch-based CNN). And Figure \ref{fig:3D result} shows that 3D-based methods have less isolated regions (false positives) than 2D-based ones. 

\textbf{(ii)} It segments NPC by fusing multi-modality MRIs with the multi-encoder network. Thus, it can learn complementary and interdependent features from different modalities of MRI for final decisions. Additionally, comparing with input-level fusion networks and decision-level fusion networks, layer-level fusion networks (including MMFNet) can effectively capture informative features from different modalities of MRI and fuse low-level features and high-level features. 

\textbf{(iii)} It uses a fusion block to fuse low-level features from different modalities of MRI and prepare these low-level features for the fusion with high-level features. Thus, it can more effectively fuse information from various sources. It also uses the self-transfer strategy to initialize the netwerk. Hereby, it can stimulate encoders to make full mining of meaningful features from modality-specific MRI. And it finally improve base multi-encoder-based network (Merging encoders' fetures) by $2.64\%$, $1.10mm$ and $11.88mm$ in $meanDSC, meanASD$ and $meanHD$.

\begin{table}[]
	\footnotesize
	\caption{NPC segmentation results of ablation experiments.}
	\centering
	\label{table:ablation experiments}
	\begin{tabular}{lccccc}
		\toprule[1.5pt]
		Method & $mean DSC$ & $mean ASD$ & $mean HD$\\
		\midrule
		\midrule
		Merging encoders' features (baseline)& $69.74\pm10.95$ & $3.17\pm3.17$ & $30.18\pm32.89$\\
		Base MMFNet &$70.15\pm10.53$ &$3.44\pm4.15$ &$24.95\pm19.66$\\
		MMFNet+multi-MLP &$70.53\pm10.20$ &$2.69\pm2.05$ &$23.85\pm21.71$\\
		MMFNet+multi-MLP+stdPool (channel-attention) &$70.73\pm11.86$ &$2.46\pm1.71$ &$20.25\pm15.04$\\
		MMFNet+multi-MLP+stdPool (spatial-attention) &$70.40\pm9.23$ &$2.74\pm2.07$ &$23.25\pm18.42$\\
		MMFNet+multi-MLP+stdPool &$71.34\pm12.38$ &$2.31\pm2.12$ &$21.06\pm17.56$\\	
		\midrule
		Base MMFNet+self-transfer &$70.76\pm12.96$ &$2.13\pm1.79$ &$\textbf{16.45}\pm8.15$\\
		MMFNet+multi-MLP+self-transfer &$71.12\pm11.41$ &$2.24\pm1.60$ &$17.68\pm9.85$\\
		MMFNet+multi-MLP+stdPool+self-transfer &$\textbf{72.38}\pm10.99$ &$\textbf{2.07}\pm2.32$ &$18.31\pm16.73$\\
		\midrule
		MMFNet+multi-MLP+stdPool+self-transfer(T1\&T2) &$70.49\pm12.98$ &$2.11\pm1.24$ &$19.23\pm20.08$\\
		MMFNet+multi-MLP+stdPool+self-transfer(T1\&CET1) &$70.42\pm12.57$ &$2.19\pm1.46$ &$18.31\pm14.89$\\
		MMFNet+multi-MLP+stdPool+self-transfer(T2\&CET1) &$68.65\pm11.83$ &$2.69\pm2.90$ &$21.89\pm20.56$\\
		\bottomrule[1.5pt]
	\end{tabular}
\end{table}

\begin{figure*}[]
	\centering	
	\subfigure[]{
		\begin{minipage}{0.135\textwidth}
			\centering
			\includegraphics[width=\textwidth]{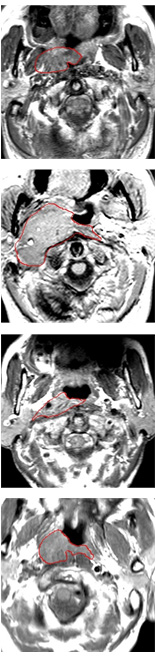}
		\end{minipage}
		\label{fig:heatmap:a}
	}	
	\hspace{-0.4cm}
	\subfigure[]{
		\begin{minipage}{0.135\textwidth}
			\centering
			\includegraphics[width=\textwidth]{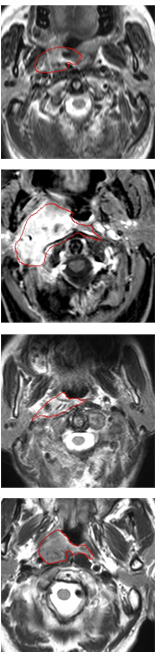}
		\end{minipage}
		\label{fig:heatmap:b}
	}	
	\hspace{-0.4cm}
	\subfigure[]{
		\begin{minipage}{0.135\textwidth}
			\centering
			\includegraphics[width=\textwidth]{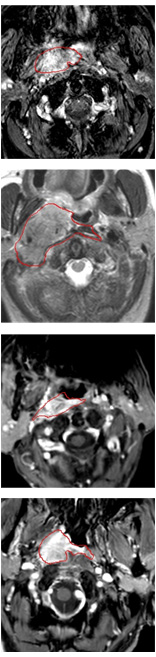}
		\end{minipage}
		\label{fig:heatmap:c}
	}	
	\hspace{-0.4cm}
	\subfigure[]{
		\begin{minipage}{0.135\textwidth}
			\centering
			\includegraphics[width=\textwidth]{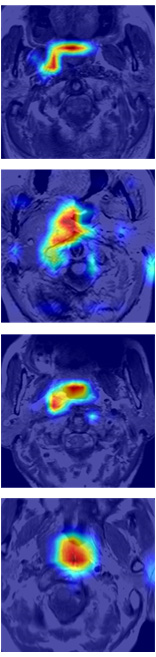}
		\end{minipage}
		\label{fig:heatmap:d}
	}	
	\hspace{-0.4cm}
	\subfigure[]{
		\begin{minipage}{0.135\textwidth}
			\centering
			\includegraphics[width=\textwidth]{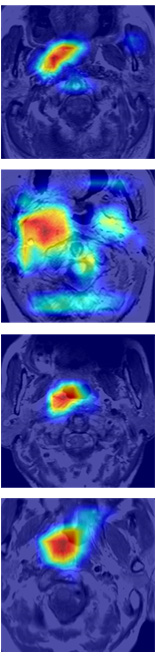}
		\end{minipage}
		\label{fig:heatmap:e}
	}	
	\hspace{-0.4cm}
	\subfigure[]{
		\begin{minipage}{0.135\textwidth}
			\centering
			\includegraphics[width=\textwidth]{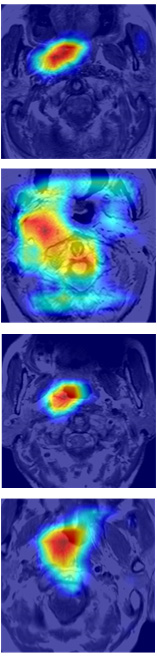}
		\end{minipage}
		\label{fig:heatmap:f}
	}	
	\hspace{-0.4cm}
	\subfigure[]{
		\begin{minipage}{0.135\textwidth}
			\centering
			\includegraphics[width=\textwidth]{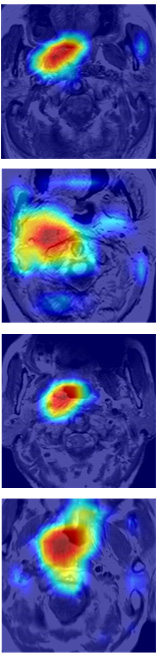}
		\end{minipage}
		\label{fig:heatmap:g}
	}	

	\caption{Examples of spatial attention coefficients. (a) T1. (b) T2. (c) CET1. (d) Spatial attention coefficients of 1/4 of the best epoch. (e) Spatial attention coefficients of 1/2 of the best epoch. (e) Spatial attention coefficients of 3/4 of the best epoch. (e) Spatial attention coefficients of the best epoch.}
	\label{fig:heatmap}
\end{figure*}
\section{Discussion}
\label{section:discussion}
In this subsection, we set extensive ablation experiments to show the effectiveness of our proposed fusion block and self-transfer. The baseline is Multi-encoder based network (Merging encoders' features), which sets individual encoders for each modality and feeds directly merging features to one single decoder.

\textbf{The design for 3D-CBAM.} The design of 3D-CBAM for base MMFNet is a simple version of 3D-CBAM, which utilizes a shared MLP in channel attention module and uses both max-pooling outputs and average-pooling outputs to obtain channel attention weights and spatial attention weights. Next, we modify single shared MLP into multiple MLPs for different global features. After that, we set several experiments to search best choice for the addition of std-pooling outputs.

According to results shown in Table \ref{table:ablation experiments}, the best design for 3D-CBAM is the one with multiple MLPs in channel attention block and with std-pooling in both channel attention and spatial attention blocks. The utilization of std-pooling outputs can provide more sufficient global information of 3D images, and the setting of multiple MLPs is suitable to deal with multiple features (outputs of std-pooling, max-pooling and average-pooling) with varied distributions.

Some examples of spatial attention coefficients are shown in Figure \ref{fig:heatmap}. We can see that, several locations may fire at the beginning of training phase and then energy will slowly build up over ROIs and reduce the attention to false positives.

\textbf{The contribution of self-transfer.} After setting several experiments to find the best design of 3D-CBAM, we implement self-transfer to these models to investigate its efficacies. We firstly train three modality-specific encoder-decoder network. Then, these pre-trained encoders aforementioned will be the initial features' extractors for several multi-modality networks with different fusion blocks.

Results in Table \ref{table:ablation experiments} show that the utilization of self-transfer can stimulate encoders to capture more meaningful features for NPC segmentation from MRI. All methods with self-transfer implemented in this paper can have better performances in evaluate metrics compared to corresponding methods without it. Therefore, self-transfer is a great strategy for multi-encoder-based network to realize NPC segmentation based on multi-modality MRI.

\textbf{The choice of MRI.} After demonstrating the effectiveness of MMFNet, we set several additional experiments to find the best choice of MRI. We set three MMFNets with two encoders to show the results of methods based on only two modalities of MRI. 

As shown in Table \ref{table:ablation experiments}, we come to the conclusion that fuse all modalities of MRI (TI, T2 and CET1) can obtain the best results. Various modalities of MRI have varied responses for different tissues. Combining all MRI to get complementary and interdependent information is meaningful for NPC segmentation.

\begin{figure*}[]
	\centering	
	\subfigure[left view]{
		\begin{minipage}{0.32\textwidth}
			\centering
			\includegraphics[width=\textwidth]{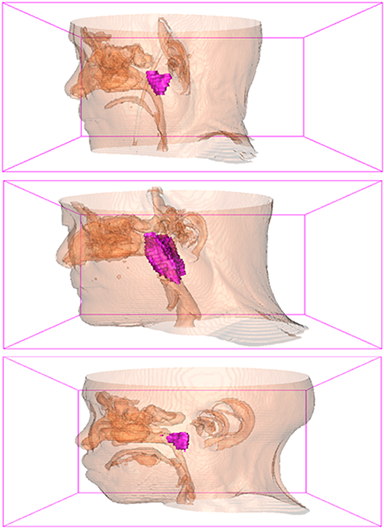}
		\end{minipage}
		\label{fig:3D_visualization:left}
	}	
	\hspace{-0.4cm}
	\subfigure[front view]{
		\begin{minipage}{0.32\textwidth}
			\centering
			\includegraphics[width=\textwidth]{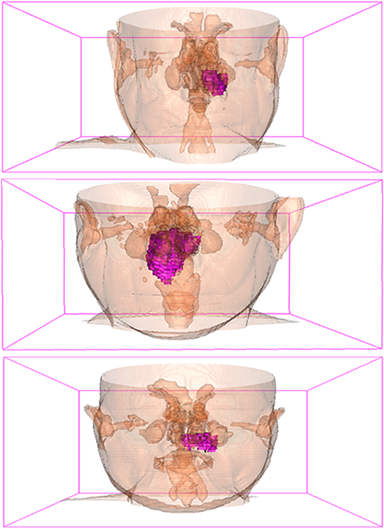}
		\end{minipage}
		\label{fig:3D_visualization:center}
	}
	\hspace{-0.4cm}
	\subfigure[right view]{
		\begin{minipage}{0.32\textwidth}
			\centering
			\includegraphics[width=\textwidth]{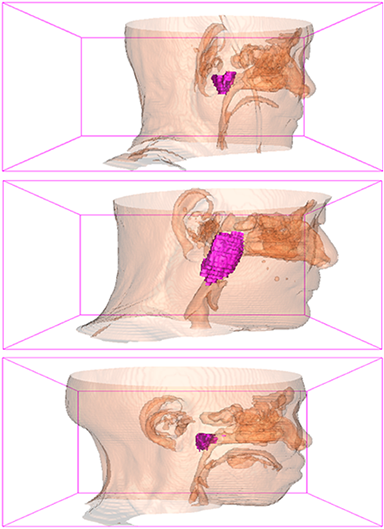}
		\end{minipage}
		\label{fig:3D_visualization:right}
	}
	
	\caption{Some visualization examples of MMFNet' predicted results, purple regions are NPC regions.}
	\label{fig:3D_visualization}
\end{figure*}

Some typical predicted results of MMFNet are visualized in Figure \ref{fig:3D_visualization}. And it is worth mentioning that our proposed network is extremely time-friendly compared to manually marking by radiologists. Specifically speaking, our proposed method only needs about 9s to realize NPC delineation of a patient, while an experienced radiologist needs 10 to 20 minutes to complete it.

\section{Conclusion}
\label{section:conclusion}
In this paper, we propose a novel multi-modality MRI fusion network (MMFNet) to segment nasopharyngeal carcinoma (NPC) based on three modalities of MRI (T1, T2 and contrast-enhanced T1). For this purpose, the backbone of the MMFNet is designed as a multi-encoder-based network, which can well learn both low-level and high-level features used implicitly for NPC segmentation in each modality of MRI. A fusion block is used in the MMFNet to effectively fuse low-level features from multi-modality MRI. It contains a 3D-CBAM and an RFBlock to recalibrate multi-source features and fuse them. A training strategy named self-transfer is proposed to effectively initialize features extractors of MMFNet. Experiments show that, the MMFNet can well segment NPC with a high accuracy and the utilization of multi-modality MRI is meaningful for the segmentation of NPC. Particularly, comparing with the existing related works, such as U-net, 3D U-net and other multi-modality-based methods, MMFNet obtain a better performance in NPC segmentation.

\textbf{Acknowledgment} This work was supported in part by NSFC of China (61375020, 61572317), Shanghai Intelligent Medicine Project (2018ZHYL0217), and SJTU Translational Medicine Cross Research Fund (ZH2018QNA05).

\section*{References}
\bibliography{mybibfile}

\begin{thebibliography}{10}
\expandafter\ifx\csname url\endcsname\relax
  \def\url#1{\texttt{#1}}\fi
\expandafter\ifx\csname urlprefix\endcsname\relax\def\urlprefix{URL }\fi
\expandafter\ifx\csname href\endcsname\relax
  \def\href#1#2{#2} \def\path#1{#1}\fi

\bibitem{mohammed2017review}
M.~A. Mohammed, M.~K.~A. Ghani, R.~I. Hamed, D.~A. Ibrahim, Review on
  nasopharyngeal carcinoma: Concepts, methods of analysis, segmentation,
  classification, prediction and impact: A review of the research literature,
  Journal of Computational Science 21 (2017) 283--298.

\bibitem{wu2018nasopharyngeal}
L.~Wu, C.~Li, L.~Pan, Nasopharyngeal carcinoma: A review of current updates,
  Experimental and therapeutic medicine 15~(4) (2018) 3687--3692.

\bibitem{razek2012mri}
A.~A. K.~A. Razek, A.~King, Mri and ct of nasopharyngeal carcinoma, American
  Journal of Roentgenology 198~(1) (2012) 11--18.

\bibitem{tatanun2010automatic}
C.~Tatanun, P.~Ritthipravat, T.~Bhongmakapat, L.~Tuntiyatorn, Automatic
  segmentation of nasopharyngeal carcinoma from ct images: region growing based
  technique, in: Signal Processing Systems (ICSPS), 2010 2nd International
  Conference on, Vol.~2, IEEE, 2010, pp. V2--537.

\bibitem{chanapai2012nasopharyngeal}
W.~Chanapai, T.~Bhongmakapat, L.~Tuntiyatorn, P.~Ritthipravat, Nasopharyngeal
  carcinoma segmentation using a region growing technique, International
  journal of computer assisted radiology and surgery 7~(3) (2012) 413--422.

\bibitem{zhou2003texture}
J.~Zhou, T.-K. Lim, V.~Chong, J.~Huang, A texture combined multispectral
  magnetic resonance imaging segmentation for nasopharyngeal carcinoma, Optical
  review 10~(5) (2003) 405--410.

\bibitem{fitton2011semi}
I.~Fitton, S.~Cornelissen, J.~C. Duppen, R.~Steenbakkers, S.~Peeters,
  F.~Hoebers, J.~H. Kaanders, P.~Nowak, C.~R. Rasch, M.~van Herk,
  Semi-automatic delineation using weighted ct-mri registered images for
  radiotherapy of nasopharyngeal cancer, Medical physics 38~(8) (2011)
  4662--4666.

\bibitem{huang2015nasopharyngeal}
K.-W. Huang, Z.-Y. Zhao, Q.~Gong, J.~Zha, L.~Chen, R.~Yang, Nasopharyngeal
  carcinoma segmentation via hmrf-em with maximum entropy, in: Engineering in
  Medicine and Biology Society (EMBC), 2015 37th Annual International
  Conference of the IEEE, IEEE, 2015, pp. 2968--2972.

\bibitem{zhou2011segmentation}
J.~Zhou, Q.~Tian, V.~Chong, W.~Xiong, W.~Huang, Z.~Wang, Segmentation of skull
  base tumors from mri using a hybrid support vector machine-based method, in:
  International Workshop on Machine Learning in Medical Imaging, Springer,
  2011, pp. 134--141.

\bibitem{huang2013region}
W.~Huang, K.~L. Chan, J.~Zhou, Region-based nasopharyngeal carcinoma lesion
  segmentation from mri using clustering-and classification-based methods with
  learning, Journal of digital imaging 26~(3) (2013) 472--482.

\bibitem{zhou2002mri}
J.~Zhou, V.~Chong, T.-K. Lim, J.~Houng, Mri tumor segmentation for
  nasopharyngeal carcinoma using knowledge-based fuzzy clustering,
  International journal of information technology 8~(2).

\bibitem{wang2018tumor}
Y.~Wang, B.~Yu, L.~Wang, C.~Zu, Y.~Luo, X.~Wu, Z.~Yang, J.~Zhou, L.~Zhou, Tumor
  segmentation via multi-modality joint dictionary learning, in: Biomedical
  Imaging (ISBI 2018), 2018 IEEE 15th International Symposium on, IEEE, 2018,
  pp. 1336--1339.

\bibitem{ma2018discriminative}
Z.~Ma, X.~Wu, S.~Sun, C.~Xia, Z.~Yang, S.~Li, J.~Zhou, A discriminative
  learning based approach for automated nasopharyngeal carcinoma segmentation
  leveraging multi-modality similarity metric learning, in: Biomedical Imaging
  (ISBI 2018), 2018 IEEE 15th International Symposium on, IEEE, 2018, pp.
  813--816.

\bibitem{ma2018automated}
Z.~Ma, X.~Wu, Q.~Song, Y.~Luo, Y.~Wang, J.~Zhou, Automated nasopharyngeal
  carcinoma segmentation in magnetic resonance images by combination of
  convolutional neural networks and graph cut, Experimental and therapeutic
  medicine 16~(3) (2018) 2511--2521.

\bibitem{wang2018automatic}
Y.~Wang, C.~Zu, G.~Hu, Y.~Luo, Z.~Ma, K.~He, X.~Wu, J.~Zhou, Automatic tumor
  segmentation with deep convolutional neural networks for radiotherapy
  applications, Neural Processing Letters (2018) 1--12.

\bibitem{popovtzer2014mri}
A.~Popovtzer, M.~Ibrahim, D.~Tatro, F.~Y. Feng, R.~K. Ten~Haken, A.~Eisbruch,
  Mri to delineate the gross tumor volume of nasopharyngeal cancers: which
  sequences and planes should be used?, Radiology and oncology 48~(3) (2014)
  323--330.

\bibitem{valindria2018multi}
V.~V. Valindria, N.~Pawlowski, M.~Rajchl, I.~Lavdas, E.~O. Aboagye, A.~G.
  Rockall, D.~Rueckert, B.~Glocker, Multi-modal learning from unpaired images:
  Application to multi-organ segmentation in ct and mri, in: Applications of
  Computer Vision (WACV), 2018 IEEE Winter Conference on, IEEE, 2018, pp.
  547--556.

\bibitem{tseng2017joint}
K.-L. Tseng, Y.-L. Lin, W.~Hsu, C.-Y. Huang, Joint sequence learning and
  cross-modality convolution for 3d biomedical segmentation, in: Computer
  Vision and Pattern Recognition (CVPR), 2017 IEEE Conference on, IEEE, 2017,
  pp. 3739--3746.

\bibitem{ma2018concatenated}
C.~Ma, G.~Luo, K.~Wang, Concatenated and connected random forests with
  multiscale patch driven active contour model for automated brain tumor
  segmentation of mr images, IEEE Transactions on Medical Imaging.

\bibitem{zhou2019review}
T.~Zhou, S.~Ruan, S.~Canu, A review: Deep learning for medical image
  segmentation using multi-modality fusion, Array (2019) 100004.

\bibitem{havaei2017brain}
M.~Havaei, A.~Davy, D.~Warde-Farley, A.~Biard, A.~Courville, Y.~Bengio, C.~Pal,
  P.-M. Jodoin, H.~Larochelle, Brain tumor segmentation with deep neural
  networks, Medical image analysis 35 (2017) 18--31.

\bibitem{kamnitsas2017efficient}
K.~Kamnitsas, C.~Ledig, V.~F. Newcombe, J.~P. Simpson, A.~D. Kane, D.~K. Menon,
  D.~Rueckert, B.~Glocker, Efficient multi-scale 3d cnn with fully connected
  crf for accurate brain lesion segmentation, Medical image analysis 36 (2017)
  61--78.

\bibitem{dolz2018hyperdense}
J.~Dolz, K.~Gopinath, J.~Yuan, H.~Lombaert, C.~Desrosiers, I.~B. Ayed,
  Hyperdense-net: A hyper-densely connected cnn for multi-modal image
  segmentation, IEEE transactions on medical imaging 38~(5) (2018) 1116--1126.

\bibitem{dolz2018ivd}
J.~Dolz, C.~Desrosiers, I.~B. Ayed, Ivd-net: Intervertebral disc localization
  and segmentation in mri with a multi-modal unet, in: International Workshop
  and Challenge on Computational Methods and Clinical Applications for Spine
  Imaging, Springer, 2018, pp. 130--143.

\bibitem{nie2016fully}
D.~Nie, L.~Wang, Y.~Gao, D.~Shen, Fully convolutional networks for
  multi-modality isointense infant brain image segmentation, in: 2016 IEEE 13Th
  international symposium on biomedical imaging (ISBI), IEEE, 2016, pp.
  1342--1345.

\bibitem{kamnitsas2017ensembles}
K.~Kamnitsas, W.~Bai, E.~Ferrante, S.~McDonagh, M.~Sinclair, N.~Pawlowski,
  M.~Rajchl, M.~Lee, B.~Kainz, D.~Rueckert, et~al., Ensembles of multiple
  models and architectures for robust brain tumour segmentation, in:
  International MICCAI Brainlesion Workshop, Springer, 2017, pp. 450--462.

\bibitem{valada2016deep}
A.~Valada, G.~L. Oliveira, T.~Brox, W.~Burgard, Deep multispectral semantic
  scene understanding of forested environments using multimodal fusion, in:
  International Symposium on Experimental Robotics, Springer, 2016, pp.
  465--477.

\bibitem{feldman2010attention}
H.~Feldman, K.~Friston, Attention, uncertainty, and free-energy, Frontiers in
  human neuroscience 4 (2010) 215.

\bibitem{itti1998model}
L.~Itti, C.~Koch, E.~Niebur, A model of saliency-based visual attention for
  rapid scene analysis, IEEE Transactions on pattern analysis and machine
  intelligence 20~(11) (1998) 1254--1259.

\bibitem{wang2017residual}
F.~Wang, M.~Jiang, C.~Qian, S.~Yang, C.~Li, H.~Zhang, X.~Wang, X.~Tang,
  Residual attention network for image classification, in: Proceedings of the
  IEEE Conference on Computer Vision and Pattern Recognition, 2017, pp.
  3156--3164.

\bibitem{hu2017squeeze}
J.~Hu, L.~Shen, G.~Sun, Squeeze-and-excitation networks, arXiv preprint
  arXiv:1709.01507 7.

\bibitem{woo2018cbam}
S.~Woo, J.~Park, J.-Y. Lee, I.~So~Kweon, Cbam: Convolutional block attention
  module, in: Proceedings of the European Conference on Computer Vision (ECCV),
  2018, pp. 3--19.

\bibitem{chen2017sca}
L.~Chen, H.~Zhang, J.~Xiao, L.~Nie, J.~Shao, W.~Liu, T.-S. Chua, Sca-cnn:
  Spatial and channel-wise attention in convolutional networks for image
  captioning, in: 2017 IEEE Conference on Computer Vision and Pattern
  Recognition (CVPR), IEEE, 2017, pp. 6298--6306.

\bibitem{simonyan2014very}
K.~Simonyan, A.~Zisserman, Very deep convolutional networks for large-scale
  image recognition, arXiv preprint arXiv:1409.1556.

\bibitem{ronneberger2015u}
O.~Ronneberger, P.~Fischer, T.~Brox, U-net: Convolutional networks for
  biomedical image segmentation, in: International Conference on Medical image
  computing and computer-assisted intervention, Springer, 2015, pp. 234--241.

\bibitem{gatys2016image}
L.~A. Gatys, A.~S. Ecker, M.~Bethge, Image style transfer using convolutional
  neural networks, in: Proceedings of the IEEE Conference on Computer Vision
  and Pattern Recognition, 2016, pp. 2414--2423.

\bibitem{szegedy2016rethinking}
C.~Szegedy, V.~Vanhoucke, S.~Ioffe, J.~Shlens, Z.~Wojna, Rethinking the
  inception architecture for computer vision, in: Proceedings of the IEEE
  conference on computer vision and pattern recognition, 2016, pp. 2818--2826.

\bibitem{milletari2016v}
F.~Milletari, N.~Navab, S.-A. Ahmadi, V-net: Fully convolutional neural
  networks for volumetric medical image segmentation, in: 3D Vision (3DV), 2016
  Fourth International Conference on, IEEE, 2016, pp. 565--571.

\bibitem{he2016deep}
K.~He, X.~Zhang, S.~Ren, J.~Sun, Deep residual learning for image recognition,
  in: Proceedings of the IEEE conference on computer vision and pattern
  recognition, 2016, pp. 770--778.

\bibitem{yosinski2014transferable}
J.~Yosinski, J.~Clune, Y.~Bengio, H.~Lipson, How transferable are features in
  deep neural networks?, in: Advances in neural information processing systems,
  2014, pp. 3320--3328.

\bibitem{yang2018denseaspp}
M.~Yang, K.~Yu, C.~Zhang, Z.~Li, K.~Yang, Denseaspp for semantic segmentation
  in street scenes, in: Proceedings of the IEEE Conference on Computer Vision
  and Pattern Recognition, 2018, pp. 3684--3692.

\bibitem{zhao2017pyramid}
H.~Zhao, J.~Shi, X.~Qi, X.~Wang, J.~Jia, Pyramid scene parsing network, in:
  IEEE Conf. on Computer Vision and Pattern Recognition (CVPR), 2017, pp.
  2881--2890.

\bibitem{huang20183d}
Y.-J. Huang, Q.~Dou, Z.-X. Wang, L.-Z. Liu, Y.~Jin, C.-F. Li, L.~Wang, H.~Chen,
  R.-H. Xu, 3d roi-aware u-net for accurate and efficient colorectal tumor
  segmentation, arXiv preprint arXiv:1806.10342.

\bibitem{kingma2014adam}
D.~P. Kingma, J.~Ba, Adam: A method for stochastic optimization, arXiv preprint
  arXiv:1412.6980.

\bibitem{cciccek20163d}
{\"O}.~{\c{C}}i{\c{c}}ek, A.~Abdulkadir, S.~S. Lienkamp, T.~Brox,
  O.~Ronneberger, 3d u-net: learning dense volumetric segmentation from sparse
  annotation, in: International Conference on Medical Image Computing and
  Computer-Assisted Intervention, Springer, 2016, pp. 424--432.

\end{thebibliography}
\end{document}